\documentclass[conference, anonymous]{IEEEtran}
\IEEEoverridecommandlockouts
\usepackage{cite}
\usepackage{amsmath,amssymb,amsfonts}
\usepackage{algorithmic}
\usepackage{graphicx}
\usepackage{textcomp}
\usepackage{xcolor}

\usepackage{times}
\usepackage{latexsym}
\usepackage[T1]{fontenc}
\usepackage[utf8]{inputenc}
\usepackage{microtype}
\usepackage{inconsolata}
\usepackage{subfigure}
\usepackage{booktabs}
\usepackage{multirow}
\usepackage{tabularx}
\usepackage{hyperref}

\def\BibTeX{{\rm B\kern-.05em{\sc i\kern-.025em b}\kern-.08em
    T\kern-.1667em\lower.7ex\hbox{E}\kern-.125emX}}
\begin{document}

\title{Building a Chinese Medical Dialogue System: Integrating Large-scale Corpora and Novel Models}

\author{
    \IEEEauthorblockN{Xinyuan Wang\textsuperscript{1}, Haozhou Li\textsuperscript{1}, Dingfang Zheng\textsuperscript{1}, Qinke Peng\textsuperscript{1}}
    \IEEEauthorblockA{\textit{Systems Engineering Institute} \\
    \textit{Xi'an Jiaotong University}, Xi'an, China \\
    \{wxy0713, lihaozhou1126, zdf0921\}@stu.xjtu.edu.cn, qkpeng@mail.xjtu.edu.cn}
}


\maketitle

\begin{abstract}
The global COVID-19 pandemic underscored major deficiencies in traditional healthcare systems, hastening the advancement of online medical services, especially in medical triage and consultation. 
However, existing studies face two main challenges. 
First, the scarcity of large-scale, publicly available, domain-specific medical datasets due to privacy concerns, with current datasets being small and limited to a few diseases, limiting the effectiveness of triage methods based on Pre-trained Language Models (PLMs). 
Second, existing methods lack medical knowledge and struggle to accurately understand professional terms and expressions in patient-doctor consultations.
To overcome these obstacles, we construct the Large-scale Chinese Medical Dialogue Corpora (LCMDC)
, thereby addressing the data shortage in this field. 
Moreover, we further propose a novel triage system that combines BERT-based supervised learning with prompt learning, as well as a GPT-based medical consultation model. 
To enhance domain knowledge acquisition, we pre-trained PLMs using our self-constructed background corpus. Experimental results on the LCMDC demonstrate the efficacy of our proposed systems.
Our corpora and codes are available at: \href{https://zenodo.org/records/13771008?token=eyJhbGciOiJIUzUxMiIsImlhdCI6MTcyNjU1NDAwMiwiZXhwIjoxNzM1NjAzMTk5fQ.eyJpZCI6IjA4Y2M0MDMyLTE0NTctNGZkZi1iYjAxLTBkZmQyYjRiNzVlZiIsImRhdGEiOnt9LCJyYW5kb20iOiI0OTExZTBhNzIyMjg5NzFhMmJmZWRhN2JmY2E2ZTljZCJ9.l7HobRPQVtt5gWBXs-2AuOsBX5fYViYkqKePsoDAvTmFYAu_1sH-2f1XwtWJJlppEAdd3C0wdWF7MbCtFLP6kA}{Corpora} and \href{https://anonymous.4open.science/r/LCMDC--Submission-4F45}{Codes}.
\end{abstract}


\section{Introduction}
The outbreak and spread of COVID-19 have placed immense pressure on global healthcare systems, exposing numerous shortcomings in traditional healthcare services \cite{sun2021covid}. In response to the urgent need for pandemic control, online medical consultations and telemedicine have become crucial for maintaining healthcare services \cite{jnr2020use}. Between July and December 2022, 58.5\% of U.S. adults used the Internet to seek health or medical information, 41.5\% communicated with a doctor or doctor’s office, and 46.1\% checked medical test results\cite{wang2023health}. According to the China Internet Network Information Center (CNNIC), as of December 2023, the number of Internet medical users in China reached 414 million. This trend reflects the growing reliance on digital platforms for healthcare.

The rise of online healthcare platforms has made medical triage and intelligent Q\&A key research areas, helping reduce the burden on healthcare systems, providing remote medical advice, and reducing unnecessary hospital visits. These technologies enhance resource efficiency, minimize cross-infection risks, and improve access to healthcare services.

However, most existing studies face two main challenges. Firstly, due to data privacy concerns, obtaining complete large-scale public medical triage and consultation datasets is difficult~\cite{complete_data}. Although some studies have collected corpora for automatic diagnosis \cite{wei2018task,lin2019enhancing,xu2019end} and dialogue generation \cite{liu2022meddg}, these datasets are limited to a few specific diseases and are small in size \cite{zeng2020meddialog}, making them inadequate for training intelligent medical diagnosis models based on pre-trained language models. Secondly, patient-doctor consultations usually utilize distinct expressions and professional terms, making it difficult for existing methods that lack domain-specific knowledge to comprehend medically pertinent information \cite{li2024sade}.

To address the limitations, we have built a Chinese medical dialogue system, including 1) the medical corpora, 2) the medical triage system, and 3) the medical consultation model.

The \textbf{L}arge-scale \textbf{C}hinese \textbf{M}edical \textbf{D}ialogue \textbf{C}orpora (\textbf{LCMDC}) extracts patient-doctor conversations from the "Quick Doctor" website \footnote{https://www.120ask.com/}. 
Our work addresses two key needs in online healthcare: task-oriented medical triage and open-ended consultation systems. We developed a coarse-grained triage dataset with over 430k patient-doctor dialogues across 14 departments and a fine-grained diagnosis dataset with 200k entries covering 120 diseases. For the consultation, we compiled over 470k question-answer pairs. Additionally, we collected disease information datasets to enrich Pre-trained Language Models (PLMs) with medical knowledge for pre-training.

Nowadays, natural language processing techniques are increasingly effective in intelligent \textbf{medical triage} and dialogue generation tasks\cite{sun2017review, xie2025transformer}. Specifically, neural network-based text classification algorithms with attention mechanisms have been successfully applied to medical triage and diagnosis\cite{wei2018task,lin2019enhancing,xu2019end}. However, existing studies often concentrate on a limited number of diseases and struggle with imbalanced label distributions in many-class medical diagnoses \cite{zeng2020meddialog}. Although some pre-trained language models, such as BERT and ERNIE, have been employed in automatic triage systems \cite{wang2021reduce, wang2022ernie}, they lack pre-training on medical data, hindering their ability to grasp domain-specific knowledge \cite{li2024sade}. To overcome these limitations, we propose a novel triage system that combines PLM-based supervised learning with prompt learning methods. We initially pre-train BERT using a self-built medical background corpus, then employ medical BERT, BiLSTM, and Dendrite Network \cite{liu2021dendrite} to represent the input questions for classification in our Triage datasets. Additionally, we incorporate prompt learning to mitigate the many-label imbalance issue, enhancing prediction accuracy for rare diseases.

In the \textbf{medical consultation model}, sequence-to-sequence models, including LSTM-based Seq2Seq methods \cite{liu2022meddg} and transformer-based PLMs like BERT-GPT \cite{zeng2020meddialog}, have been widely used, and achieved promising results. However, traditional LSTM models, which generate output solely based on input sequences, struggle with insufficient information utilization due to their gating mechanisms. Similarly, conventional PLMs often fail to effectively comprehend medical information within patient-doctor conversations. Additionally, these methods typically lack supplementary information for enhancement, resulting in poor performance in generating patient-doctor dialogues. To address these challenges, we re-train GPT-2 using our collected medical background corpus to internalize domain-specific knowledge. We also construct a domain knowledge graph and extract relevant content about each user’s question as supplementary input information to enhance the model. Finally, we train our model using our Medical Consultation Dataset. 

The contributions of this paper are three-fold:

\begin{itemize}
  \item Collected and integrated over millions of samples to form three Large-scale Chinese Medical Corpus. To support medical domain language models and lay the foundation for automatic triage and medical consulting systems.
  \item Proposed an information fusion-based classification algorithm and a prompt learning classification algorithm to improve intelligent triage accuracy, especially for rare diseases, achieving a 5\% improvement in accuracy.
  \item Developed a medical consultation model including knowledge incorporation and input supplementation, improving performance across multiple evaluation metrics.
\end{itemize}

\section{Related Work}
\label{sec:relatedwork}
The development of AI-driven medical dialogue systems benefits from advancements in biomedical research and feature engineering techniques. Studies on bioinformatics and medicine~\cite{dcai_survey, liu2024pth,wang2022successful,liu2024calorie, xu2022wnt4,leng2022inflammation} provide valuable domain knowledge that can enhance medical AI applications. Additionally, effective feature selection and transformation techniques are crucial for optimizing model performance. Feature selection methods~\cite{ying2024revolutionizing,wang2024knockoff,ying2024feature,gong2024neuro} help identify the most informative clinical attributes, while feature transformation approaches~\cite{gong2024evolutionary,ying2023self,ying2024unsupervised,ying2024topology,hu2024reinforcement} improve data representation, enabling more accurate and robust medical triage and consulting systems. Furthermore, large language models (LLMs) have significantly advanced research in medical AI by improving the understanding and processing of complex biomedical texts~\cite{wang2024llm,li2023sehf,wang2022hierarchal}. Pre-trained models with domain-specific fine-tuning enable more effective medical dialogue generation, question answering, and clinical decision support.

\subsection{Dataset}
CBLUE \cite{zhang2021cblue} is a comprehensive Chinese biomedical language understanding benchmark designed to advance research in biomedical NLP for the Chinese language. MedDG \cite{liu2022meddg} is a large-scale dataset aimed at developing conversational agents for primary clinical advice, especially valuable during the COVID-19 pandemic. MedDialog \cite{zeng2020meddialog} is a comprehensive dataset for training and evaluating medical dialogue systems, enhancing remote healthcare services. MedQuAD \cite{ben2019question} is a significant resource for developing AI models in medical question answering, providing a robust foundation for understanding and responding to medical inquiries. There are some other datasets like event recognition~\cite{deng20222event}.

\subsection{Triage System}
IntelTriage \cite{billis2019towards} is an intelligent triage system for Emergency Departments in Greece, dynamically prioritizing patients and monitoring their vital signs using wearable biosensors. The Babylon AI-powered Triage and Diagnostic System \cite{razzaki2018comparative} demonstrated diagnostic accuracy comparable to human doctors and provided safer triage recommendations. MESTRIMAN \cite{sierra1993mestriman} is an expert system designed to support medical assistance during catastrophes by coordinating resources and enhancing disaster management. The SADE model \cite{li2024sade} addresses challenges in medical triage and diagnosis by using dual encoders to represent patient consultations and doctor diagnoses.

\subsection{Question Answer System}
This research \cite{abdallah2020automated} automates the generation of qualified medical answers using an end-to-end RNN model trained on data from online health services. Another system \cite{jiang2021research} integrates medical knowledge, knowledge graphs, and QA systems, using crawler technology and rule-based algorithms for classification and querying. An end-to-end character-level multi-scale CNN framework \cite{zhang2017chinese} is introduced for Chinese medical question-answer matching. Med-PaLM 2 \cite{singhal2023towards} leverages large language model improvements and medical domain fine-tuning to enhance performance on medical QA datasets.
\section{Chinese Medical Corpus}
\label{corpus}
To address the scarcity of large-scale corpora and the constraints of current datasets, we collected millions of doctor-patient conversations to construct the \textbf{L}arge-scale \textbf{C}hinese \textbf{M}edical \textbf{D}ialogue \textbf{C}orpora (\textbf{LCMDC}). This collection is designed to enhance medical triage and dialogue generation and includes three components: 1) a Coarse-grained Triage dataset, 2) a Fine-grained Diagnosis dataset, and 3) a Medical Consultation dataset.

\subsection{Data Collection}
To make the model more adaptable to conversational and ambiguous medical descriptions, we claw the original data from the online medical consultation platform named "Quick Doctor".
Here are several reasons:
\begin{itemize}
    \item Apart from sensitive information, the patient consultations and doctor responses are public, reducing the difficulty of data collection.
    \item The portal boasts an extensive dataset, containing over 100 million entries, which adequately meets our requirements.
    \item The portal provides comprehensive information, including user characteristics like gender and age, and detailed classifications of medical departments across four levels, facilitating robust data statistics and medical department recommendations.
\end{itemize}

\begin{figure}[htbp]
\centering
\subfigure[Age Distribution]{
    \includegraphics[width=0.22\textwidth]{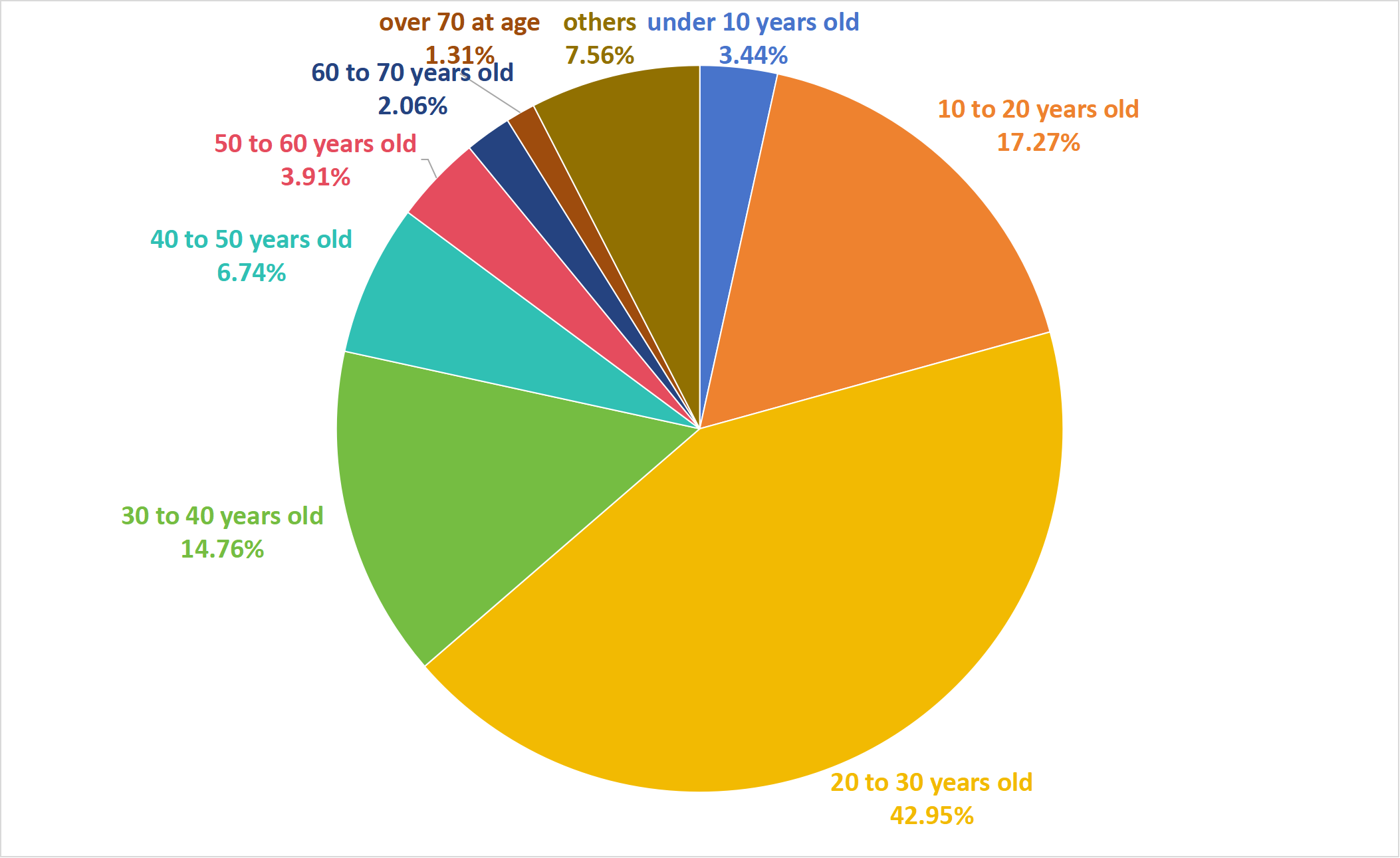}
    \label{fig:dataset_subfigure1}
}
\hfill
\subfigure[Gender Distribution]{
    \includegraphics[width=0.22\textwidth]{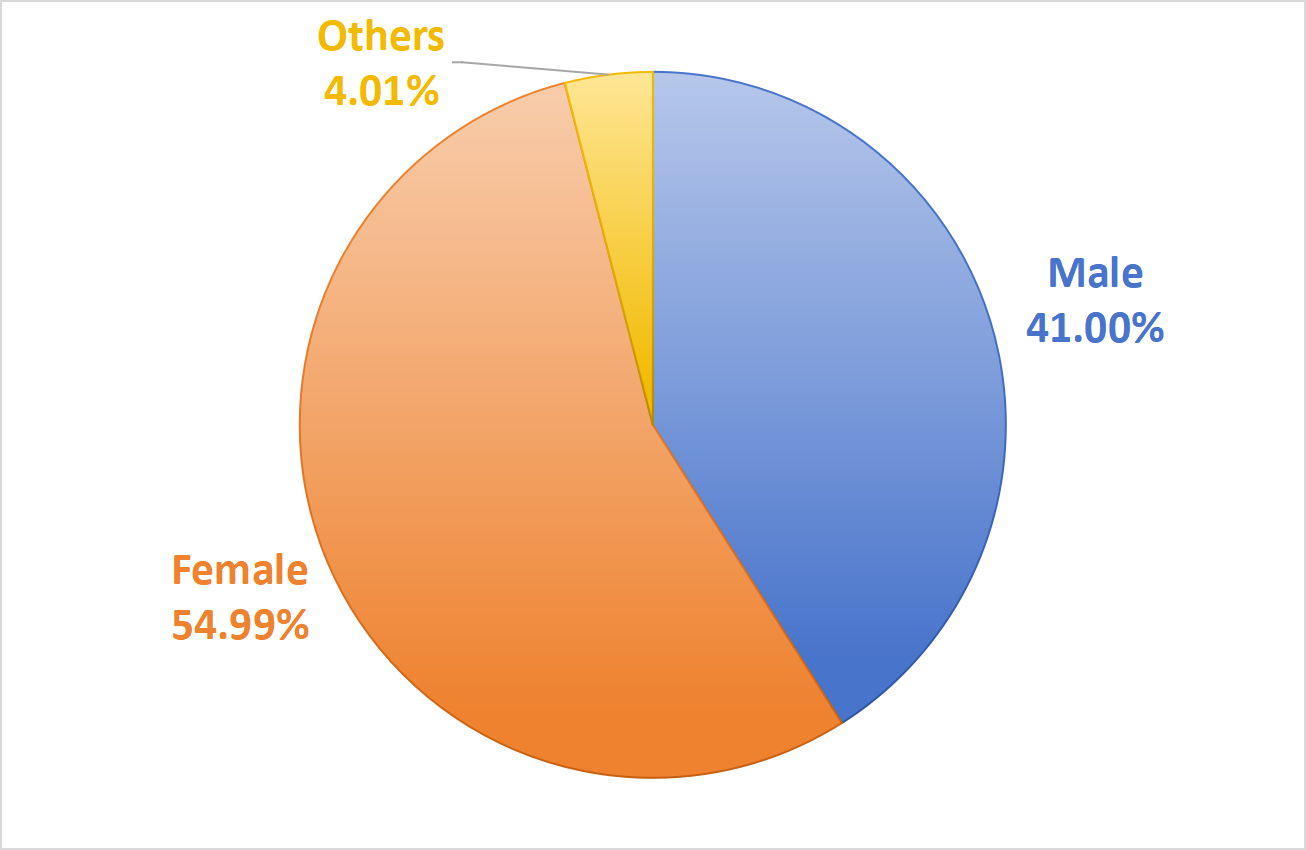}
    \label{fig:dataset_subfigure2}
}
\caption{Age and Gender Distribution in our Corpus.}
\label{fig: datasets}
\vspace{-0.2cm}
\end{figure}

We collect millions of entries containing doctor responses and conduct statistical analysis and visualization based on different labels to better understand the original data. Our analysis, illustrated in Figure \ref{fig: datasets}, explores the gender and age of users. The results show that female users significantly outnumber male users. Predominantly, young individuals aged 10-30, who are more likely to incorporate the Internet into their daily routines, made the inquiries.

\subsection{Intelligent Triage Dataset}
For automatic medical triage and diagnosis, we construct two datasets of different granularities for experimental analysis. The triage process is conceptualized as a text classification task, where patient consultations serve as input, and department classifications function as labels.

We organize the original corpus into two datasets after data preprocessing: 1) a Coarse-grained Triage dataset with primary department labels for higher accuracy in quickly identifying potential conditions and directing patients to the appropriate department, and 2) a Fine-grained Diagnosis dataset with tertiary department labels for more precise guidance, thereby helping users access detailed medical knowledge and improving query efficiency. The Coarse-grained dataset has over 430k samples with 14 categories, while the Fine-grained dataset has over 190k samples with 120 categories, as shown in Table \ref{tab:clsdataset}. Based on these datasets, we further propose an intelligent triage algorithm based on supervised learning and prompt-based learning and then evaluate its effectiveness.

\vspace{-0.4cm}
\begin{table}[htbp]
\caption{Statistic of Intelligent Triage Datasets}
\vspace{-0.6cm}
\begin{center}
\resizebox{\linewidth}{!}{
\begin{tabular}{lcc}
\toprule
\textbf{Dataset Name} & \textbf{Coarse-grained} & \textbf{Fine-grained} \\
\midrule
Number of Data & 439,630 & 199,600 \\
Training Set & 373,686 & 169,660 \\
Test Set & 65,944 & 29,940 \\
Average Sentence Length & 57.77 & 58.30 \\
Number of Categories & 14 & 120 \\
\bottomrule
\end{tabular}
\label{tab:clsdataset}
}
\end{center}
\vspace{-0.4cm}
\end{table}

\subsection{Medical Consultation Dataset}
For the medical dialogue generation, we extract the Question-Answer sample from the original corpus to create the Medical Consultation Dataset, which can offer several advantages: 1) Major medical Q\&A portals have a large user base with comprehensive questions, providing a wide range of input data for the model to learn from; 2) High-quality doctors registered on these sites provide accurate responses, providing high-quality domain knowledge for training a professional model; 3) The large data volume allows extensive training to improve model performance from various perspectives.

\vspace{-0.3cm}
\begin{table}[htbp]
\caption{Statistic of Medical Consulting Dataset}
\vspace{-0.4cm}
\begin{center}
\resizebox{0.8\linewidth}{!}{
\begin{tabular}{lcc}
\toprule
\textbf{Dataset} & \textbf{Item} & \textbf{Quantity} \\
\midrule
Q\&A Data & Total Number of Q\&A & 472,418 \\
\midrule
\multirow{3}{*}{Training Set} & Number of Entries & 467,418 \\
 & Average Question Length & 57.86 \\
 & Average Answer Length & 92.57 \\
\midrule
\multirow{3}{*}{Test Set} & Number of Entries & 5,000 \\
 & Average Question Length & 57.88 \\
 & Average Answer Length & 92.54 \\
\bottomrule
\end{tabular}
\label{tab:qadataset}
}
\end{center}
\vspace{-0.3cm}
\end{table}

Specifically, each sample in this dataset comprises a patient consultation and a doctor's response, serving respectively as the model's input and prediction target. We conducted data cleaning to remove entries that lacked either questions or answers and eliminated entries where the questions and responses were shorter than 10 characters, thereby enriching the training data with more substantive content. After data preprocessing, the Medical Consultation Dataset contains a total of 472,418 samples, with an average question length of 57.86 characters and an average response length of 92.57 characters. The statistical results are presented in Table \ref{tab:qadataset}.

\section{Intelligent Triage System}
We propose two text classification-based methods to evaluate intelligent triage datasets, using coarse-grained and fine-grained data to provide varying levels of consultation suggestions. We re-train language models on medical text data to improve their performance on Chinese medical texts and develop a supervised text classification model based on these domain-adapted models. For categories with limited data, a prompt-based text classification method leverages pre-trained language models' knowledge, enhancing performance on small datasets.

\subsection{Problem Definition}
The intelligent medical triage task is modeled as a text classification problem, processing user queries to determine the appropriate department for consultation.
\vspace{-0.2cm}
\begin{equation}
p\left\{ {c_{1},~c_{2},\ldots,c_{k_{l}}} \right\} = TextCls\left( {t_{1},t_{2},\ldots,t_{k_{w}}} \right),
\end{equation}

\subsection{Supervised Learning Triage System}
\begin{figure}[htbp]
\centering
\includegraphics[width=0.4\textwidth]{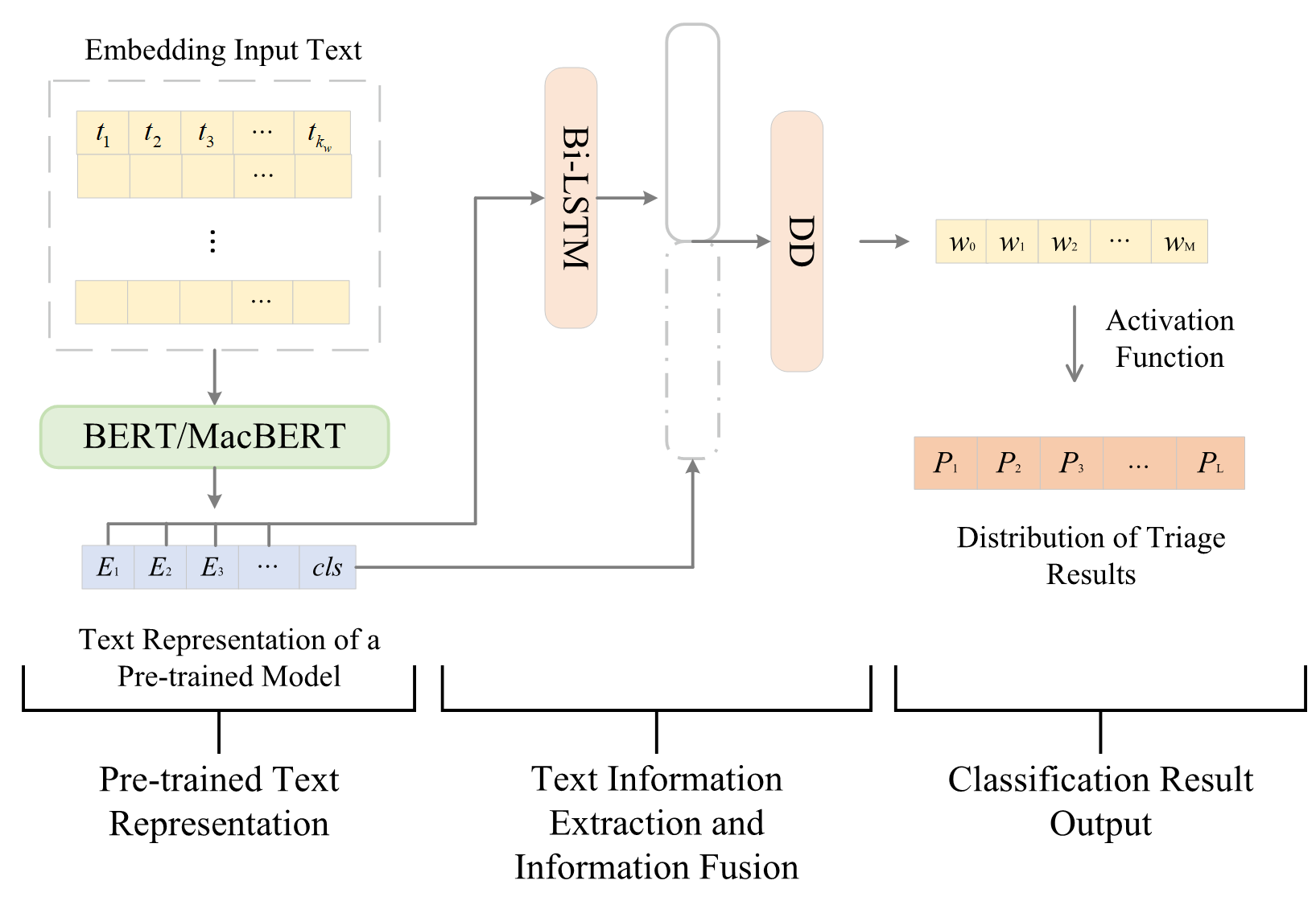}
\caption{The Proposed Supervised Learning Classification Method.}
\label{fig:supervisedcls}
\vspace{-0.3cm}
\end{figure}

First, we propose a traditional supervised learning-based text classification method. The proposed network framework is shown in Figure \ref{fig:supervisedcls}. The input is the user query text, and the output is the classification distribution of the text. First, we use a pre-trained language model (e.g., BERT) to obtain the CLS vector of the text, which is the classification vector inherent in BERT. Simultaneously, we extract the token representation vectors encoded by BERT and use a bidirectional LSTM network to capture temporal information. The CLS vector and the output vectors from the LSTM layer are then fused and processed using a Dendritic Network for information integration. Finally, a Single-Layer Perceptron and a Softmax function are used to obtain the classification distribution.

The model consists of four steps: 
The first step is domain-specific knowledge learning based on pre-training. To make the pre-trained language model used in this model more suitable for the medical domain, we further pre-trained the general pre-trained language model using the collected medical encyclopedia data and medical consulting data. Self-supervised learning was performed using the MLM pre-training task, enabling the model to better learn medical knowledge and colloquial expressions. The pre-training process employed the same parameters as BERT, selecting 15\% of the total words as masked tokens. The training objective of the MLM task is to minimize the following negative log-likelihood loss function: 

\begin{equation}
L = - {\sum\limits_{\hat{x} \in m{(x)}}{\log{P\left( \left. \hat{x} \right| \right.}}}\left. X_{\backslash m{(x)}} \right)
\end{equation}

Based on the pre-trained language model with medical text supplementation, we represent the text by obtaining the classification vector inherent to the pre-trained language model and the text token representation vector encoded by the pre-trained language model. 

\begin{equation}
\mathbf{X}_{i}^{emb} = BERTemb\left( T_{i} \right)
\end{equation}

The embedding vectors are then computed by the multi-head self-attention mechanism in BERT to get the representation vectors and the CLS vector.

\begin{equation}
\mathbf{S}_{i}^{l,j} = softmax\left( \frac{\mathbf{Q}_{i}^{l,j}*{\mathbf{K}_{i}^{l,j}}^{T}}{\sqrt{k_{h}}} \right)*\mathbf{V}_{i}^{l,j}~,~
\end{equation}

where $\mathbf{S}_{i}^{l,j} \in \mathbb{R}^{k_{m} \times k_{h}}$ are representations with attention.

\begin{equation}
{\mathbf{O}_{i}^{l}}_{(1)} = LayerNorm\left( {\mathbf{X}_{i}^{l - 1} + {\mathbf{S}_{i}^{l}}_{(2)}} \right)~,
\end{equation}

where $~{\mathbf{O}_{i}^{l}}_{(1)} \in \mathbb{R}^{k_{m} \times k_{h}}$ are updated representations.

Next, the temporal information is extracted using a bidirectional LSTM network 

\begin{equation}
\begin{aligned}
{\overset{\rightarrow}{\mathbf{H}}}_{i} &= \overset{\rightarrow}{LSTM}\left( \mathbf{P}_{i} \right)\\
{\overset{\leftarrow}{\mathbf{H}}}_{i} &= \overset{\leftarrow}{LSTM}\left( \mathbf{P}_{i} \right)\\
\mathbf{H}_{i} &= \left\lbrack {~{\overset{\rightarrow}{\mathbf{H}}}_{i},~{\overset{\leftarrow}{\mathbf{H}}}_{i}} \right\rbrack,~\mathbf{H}_{i} \in \mathbb{R}^{2 \bullet k_{h}}
\end{aligned}
\end{equation}

After obtaining the representation \( \mathbf{H}_i \) summarized by the LSTM network, we fuse it with the BERT output classification vector \([ \text{Cls} ]_i\) to form a new representation vector \( \mathbf{M}_i \):

\begin{equation}
\mathbf{M}_i = [ \mathbf{H}_i ; \text{Cls}_i ]
\end{equation}

This vector is then input into the DD layer for processing, which reduces the probability of overfitting and improves the model's robustness.

\begin{equation}
\mathbf{D}_{i} = \mathbf{W}_{\mathbf{d}}*\left( {\mathbf{M}_{i} \bullet \mathbf{M}_{i}} \right),\mathbf{W}_{\mathbf{d}} \in \mathbb{R}^{{({3 \bullet k_{h}})} \times k_{h}}
\end{equation}

Finally, the decision is made through a dense layer:

\begin{equation}
\mathbf{C}_{p} = Softmax\left( {\mathbf{D}_{i}*\mathbf{W}_{F} + \mathbf{b}_{F}} \right),
\end{equation}
where $\mathbf{W}_{F} \in \mathbb{R}^{k_{h} \times k_{l}}$ and $\mathbf{b}_{F} \in \mathbb{R}^{k_{l}}$.
And then the loss function used here is:

\begin{equation}
loss = - {\sum\limits_{i = 1}^{k_{l}}{c_{i}{\mathit{\log}\left( p_{i} \right)}}}
\end{equation}

\subsection{Prompt Learning Triage System}

\begin{figure}[htbp]
\centering
\includegraphics[width=0.4\textwidth]{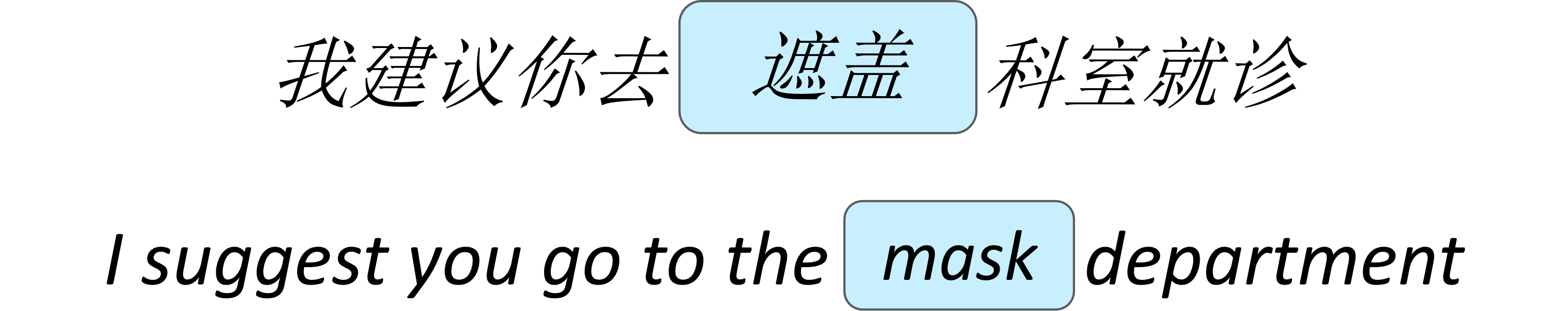}
\caption{The Proposed Prompt Structure for Classification.}
\label{fig:promptcls}
\end{figure}

Both granularity-level classification datasets suffer from data imbalance issues. For categories with less data, the performance of conventional text classification methods encounters certain bottlenecks. To address this, we propose a prompt-based classification method that leverages the knowledge embedded in pre-trained language models tailored to the medical domain to improve classification accuracy. In this prompt-based classification method, the label is a list of text, and the classification process involves predicting the content of this text, thus achieving classification. This approach establishes a relationship between the input sentence and the label name. It effectively utilizes the knowledge within pre-trained language models and performs well on datasets with few samples, making it suitable for classifying rare diseases with limited samples.
Specifically, for each user input text, we add a prompt sentence with a masked word, shown in Figure \ref{fig:promptcls}.

In this way, the classification task transforms into predicting the masked words. Interestingly, this training method aligns with the MLM task in BERT's pre-training process, allowing us to leverage the knowledge acquired during pre-training for fill-in-the-blank tasks. This accelerates the training process and is suitable for categories with fewer training samples.

\section{Medical Consultation System}
For the medical consultation dataset, we developed a Chinese medical dialogue system using a generative language model to simulate interactions with doctors. The system incorporates medical knowledge and supplements inputs, with user questions as input text and doctor's responses as output text for training.

\subsection{Problem Definition}
We mathematically model building a consultation system as a text-generation task. Specifically, based on the input question \( Q = (q_1, q_2, \ldots, q_M) \), we use a knowledge graph for supplementary input, denoted as \( I = Graph(Q) = (i_1, i_2, \ldots, i_L) \). The text generation model produces the response sequence \( A = (a_1, a_2, \ldots, a_N) \) as follows:

\begin{equation}
\begin{aligned}
\mathbf{A} &= TextGen\left( \mathbf{Q}, \mathbf{I} \right) \\
           &= TextGen\left( q_{1}, q_{2}, \ldots, q_{M}, i_{1}, i_{2}, \ldots, i_{L} \right)
\end{aligned}
\end{equation}

\subsection{Framework Overview}
Based on the GPT-2 model, we re-train the model using the collected medical encyclopedia text data to internalize knowledge, making it more suitable for tasks in the medical domain. Next, we construct a medical knowledge graph and use a search-and-match approach to find relevant knowledge to each question, providing external information as supplementary input text. Then, we train the model using dialogue data.

\begin{figure}[htbp]
\centering
\includegraphics[width=0.4\textwidth]{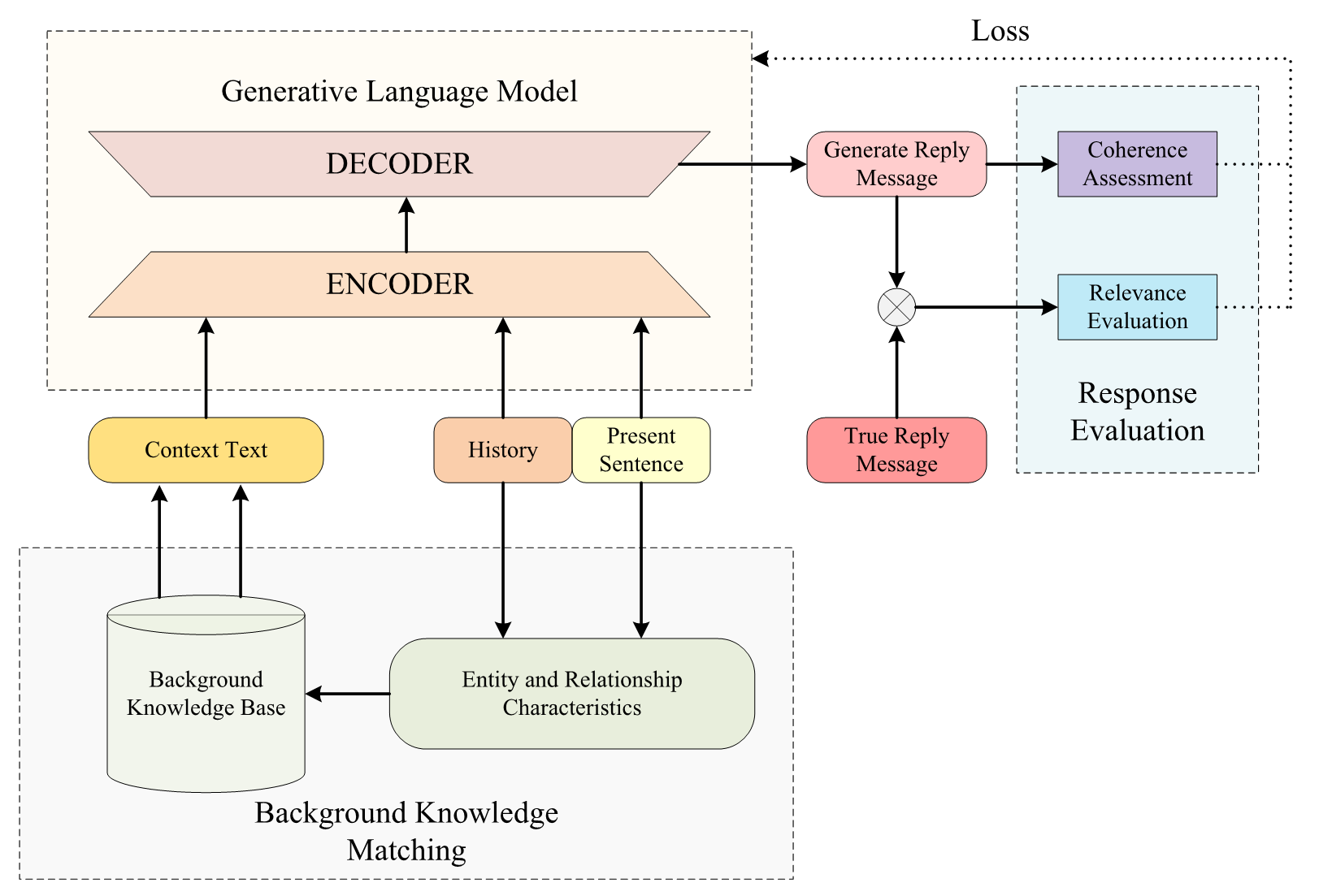}
\vspace{-0.3cm}
\caption{The Medical Dialogue System Framework.}
\label{fig:gen}
\vspace{-0.3cm}
\end{figure}

According to the above process, we construct the domain-specific medical consultation system, as shown in Figure \ref{fig:gen}.

\subsection{Knowledge Injection Answer Generation}
Knowledge injection allows the model to have rich prior knowledge, which involves re-training the large language model using the collected basic knowledge text data. After such training, the originally broad language model can focus more on the medical field, incorporating this medical knowledge to support subsequent medical consultation.

In the re-training stage, we view a sentence as a token sequence:
$Sentence = \left( {{token}_{1}, {token}_{2}, \ldots, {token}_{N}} \right)$. The objective is to maximize the following objective function:

\vspace{-0.3cm}
\begin{equation}
\scriptsize
L\left( Sentence \right) = 
\sum\limits_{n = K}^{N} 
p\left( 
{token}_{n} \,\middle|\, 
{token}_{n - K}, \ldots, 
{token}_{n - 1}; \theta 
\right)
\end{equation}
where \(\theta\) represents the parameters in the GPT-2 model that need to be fine-tuned, and encompass knowledge.

Knowledge injection is crucial because large models are trained on diverse datasets, making them broad but not sufficiently focused on specific domains. Internalizing knowledge helps the model concentrate more on the required domain and incorporate more domain knowledge.
Next is the input supplementation process, which involves constructing a knowledge graph and searching for relevant text information based on the input text. This information, combined with the input question text, serves as new input for external knowledge supplementation. This process simulates how humans search for relevant knowledge based on the question and respond using both the question and related knowledge.
The subsequent stage involves training the model with domain-specific data. This stage aims to teach the model the dialogue patterns and generate responses based on input data. This corresponds to the fine-tuning phase of the pre-trained language model, enabling the model to simulate human doctors in responding.

\section{Experiments}
We build the intelligent triage system and medical consultation system and test them on the collected datasets.

\subsection{Experiment Setup}

\subsubsection{Datasets} 
We used the datasets shown in Section \ref{corpus}, including the intelligent triage and consultant datasets.

\subsubsection{Baselines}
We employ the mechanism learning baselines, neural network baselines, and pre-trained language models, shown in Table~\ref{tab:combined_results} and Table~\ref{tab:comparison_text_generation}.

\subsubsection{Evaluation Metrics}
For the \textbf{triage system}, we utilize accuracy, precision, recall, and F-1 score.
For the \textbf{consultation system}, we employ n-gram overlap, distance-based, vector similarity, and sentence semantic similarity measurements.

\subsubsection{Hyper Parameters}

\textbf{Continuous pre-training}: Max length 128, mask 15\% tokens in MLM.
\textbf{Supervised training}: Embedding dim 768, LSTM hidden size 1024, 50 epochs. PTM lr 5e-5, others 2e-4.
\textbf{Prompt learning}: 20 epochs, lr 2e-5.
\textbf{Consultation system}: Base model GPT-2, token embedding dim 768, vocab length 50257. Pre-train 10 epochs with background knowledge, fine-tune 20 epochs, lr 2.6e-5.

\subsubsection{Experimental Environment}
All experiments were conducted on Ubuntu 22.04.3 LTS OS, with the framework of Python 3.11.5 and PyTorch 2.0.1. All data are computed on 4 NVIDIA GeForce RTX 2080 Ti GPUs, each featuring 12 GB of memory with CUDA version 12.2.

\subsection{Supervised Triage System Experiment}

\subsubsection{Overall Performance}
For the supervised learning classification method, we employ different baselines including machine learning methods, neural network methods, and pre-trained models. The results of different-grained datasets are shown in Table \ref{tab:combined_results}.

\begin{table*}[htbp]
\caption{Experimental Results on Different Granularity Datasets}
\vspace{-0.5cm}
\begin{center}
\resizebox{\textwidth}{!}{
\begin{tabular}{ll||cccc||cccc}
\toprule
\multicolumn{2}{c||}{} & \multicolumn{4}{c||}{\textbf{Coarse-grained Dataset Results}} & \multicolumn{4}{c}{\textbf{Fine-grained Dataset Results}} \\
\midrule
\textbf{Classification Method} & \textbf{Model} & \textbf{Accuracy} & \textbf{F1 Score} & \textbf{Precision} & \textbf{Recall} & \textbf{Accuracy} & \textbf{F1 Score} & \textbf{Precision} & \textbf{Recall} \\
\midrule
\multirow{2}{*}{Machine Learning} & SVM & 57.27\% & 44.27\% & 46.67\% & 42.10\% & 62.68\% & 53.83\% & 56.75\% & 51.19\% \\
 & Decision Tree & 61.24\% & 48.86\% & 52.47\% & 45.71\% & 59.68\% & 45.61\% & 45.49\% & 45.74\% \\
\midrule
\multirow{5}{*}{Neural Network} & TextCNN~\cite{kim2014convolutional} & 77.07\% & 54.36\% & 56.83\% & 52.10\% & 76.38\% & 61.29\% & 66.92\% & 56.54\% \\
 & RCNN~\cite{lai2015recurrent} & 71.56\% & 51.40\% & 53.76\% & 49.24\% & 73.84\% & 59.86\% & 61.10\% & 58.67\% \\
 & LSTM & 73.38\% & 53.02\% & 55.37\% & 50.86\% & 71.59\% & 59.80\% & 66.91\% & 54.08\% \\
 & Self Attention & 64.52\% & 52.17\% & 54.22\% & 50.27\% & 68.08\% & 60.17\% & 64.90\% & 56.08\% \\
 & Capsule~\cite{sabour2017dynamic} & 62.70\% & 51.47\% & 53.01\% & 50.01\% & 65.98\% & 52.23\% & 57.88\% & 47.59\% \\
\midrule
\multirow{3}{*}{Pre-trained Model} & BERT & 79.60\% & 59.92\% & 62.51\% & 58.79\% & 80.53\% & 65.65\% & 70.64\% & 65.24\% \\
 & RoBERTa~\cite{liu2019roberta} & 77.59\% & 55.69\% & 60.23\% & 53.98\% & 77.31\% & 55.74\% & 61.37\% & 53.46\% \\
 & MacBERT~\cite{cui2020revisiting} & 80.17\% & 59.68\% & \textbf{66.56\%} & 59.84\% & 80.73\% & 65.23\% & 71.13\% & 64.88\% \\
\midrule
\multirow{2}{*}{Experimental Algorithm} & No PT Benchmark & 81.34\% & 63.20\% & 65.05\% & 62.76\% & 82.47\% & 72.01\% & 75.60\% & 69.78\% \\
 & Our Method & \textbf{82.05\%} & \textbf{65.60\%} & 66.54\% & \textbf{65.72\%} & \textbf{83.31\%} & \textbf{74.12\%} & \textbf{77.31\%} & \textbf{72.30\%} \\
\bottomrule
\end{tabular}
}
\end{center}
\label{tab:combined_results}
\vspace{-0.5cm}
\end{table*}

According to the experimental results, the proposed supervised learning-based text classification algorithm demonstrates better classification performance compared to machine learning and deep neural network-based text classification algorithms. It shows improvements in metrics such as accuracy and F1-score, indicating the effectiveness of the proposed algorithm. Additionally, compared to commonly used pre-trained language models, the proposed classification algorithm leverages more information and effectively integrates this information, leading to enhanced classification performance.

When comparing the classification results on the coarse-grained and fine-grained datasets, it is found that the classification performance on the fine-grained dataset, which has more categories, is actually better. This issue can be attributed to the nature of the datasets themselves. Both granularity-level datasets are derived from the same batch of collected data, with the coarse-grained dataset containing over 430,000 entries, while the fine-grained dataset contains only about 190,000 entries. Since the fine-grained dataset uses tertiary department or disease labels, these labels provided by human doctors are very precise. However, the coarse-grained dataset only requires primary labels, leading to many texts in the coarse-grained dataset being ambiguously classified, making it difficult even for human doctors to provide suitably detailed labels. This results in better classification outcomes on the fine-grained dataset compared to the coarse-grained dataset.

\subsubsection{Ablation study}
\begin{figure}[htbp]
\centering
\subfigure[Coarse-grained]{
    \includegraphics[width=0.22\textwidth]{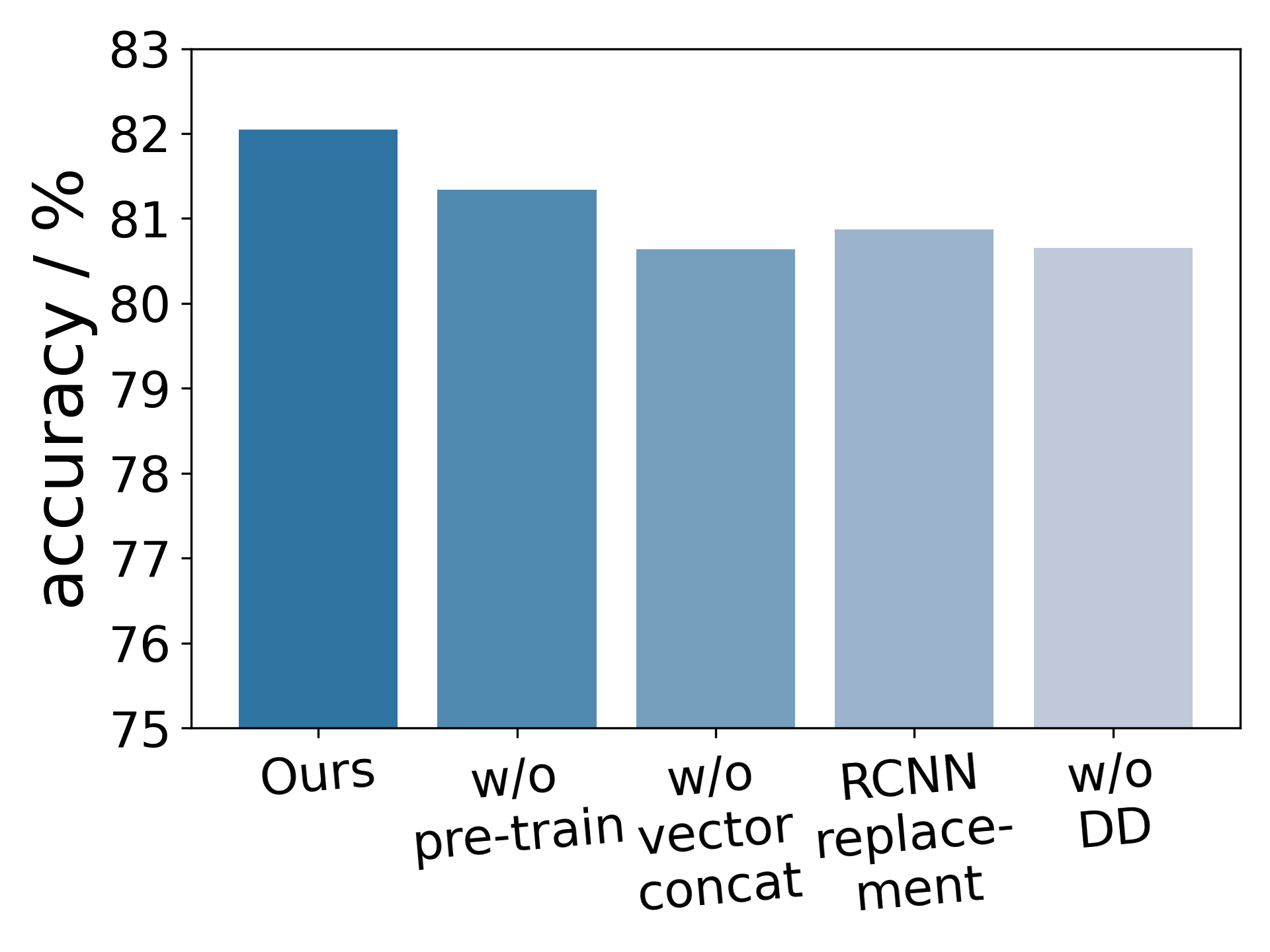}
    \label{fig:ablation_subfigure1}
}
\hfill
\subfigure[Fine-grained]{
    \includegraphics[width=0.22\textwidth]{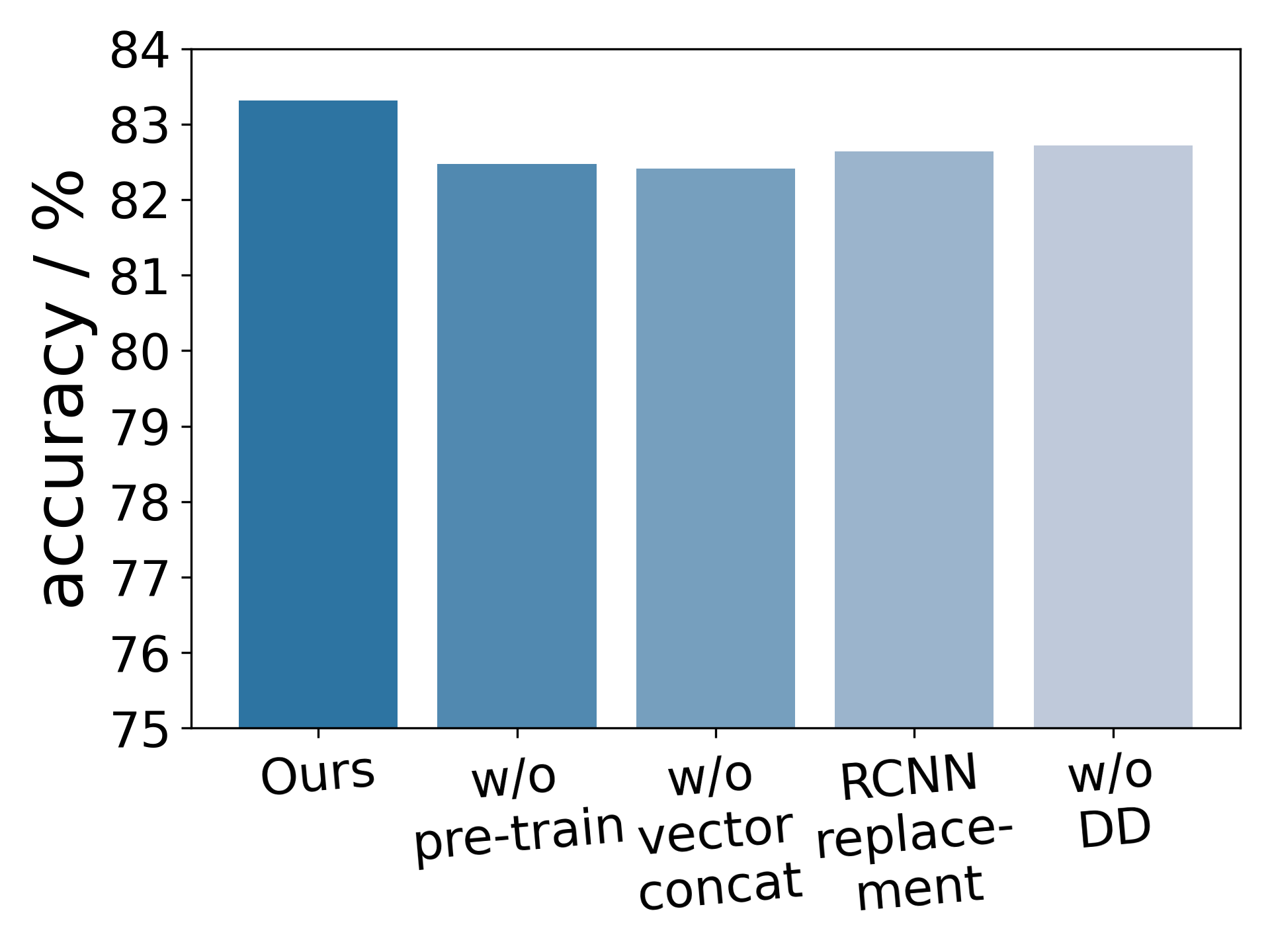}
    \label{fig:ablation_subfigure2}
}
\vspace{-0.2cm}
\caption{Ablation Results of Supervised Classification Method.}
\vspace{-0.2cm}
\label{fig: ablation}
\end{figure}

When any part of the proposed classification algorithm is removed or replaced with another structure, the classification accuracy of the new model decreases, as shown in Figure \ref{fig: ablation}. This indirectly highlights the importance of each component of the proposed model.

\vspace{-0.3cm}

\subsubsection{Results of Domain Information Injection}
\begin{figure}[htbp]
\centering
\subfigure[Coarse-grained]{
    \includegraphics[width=0.22\textwidth]{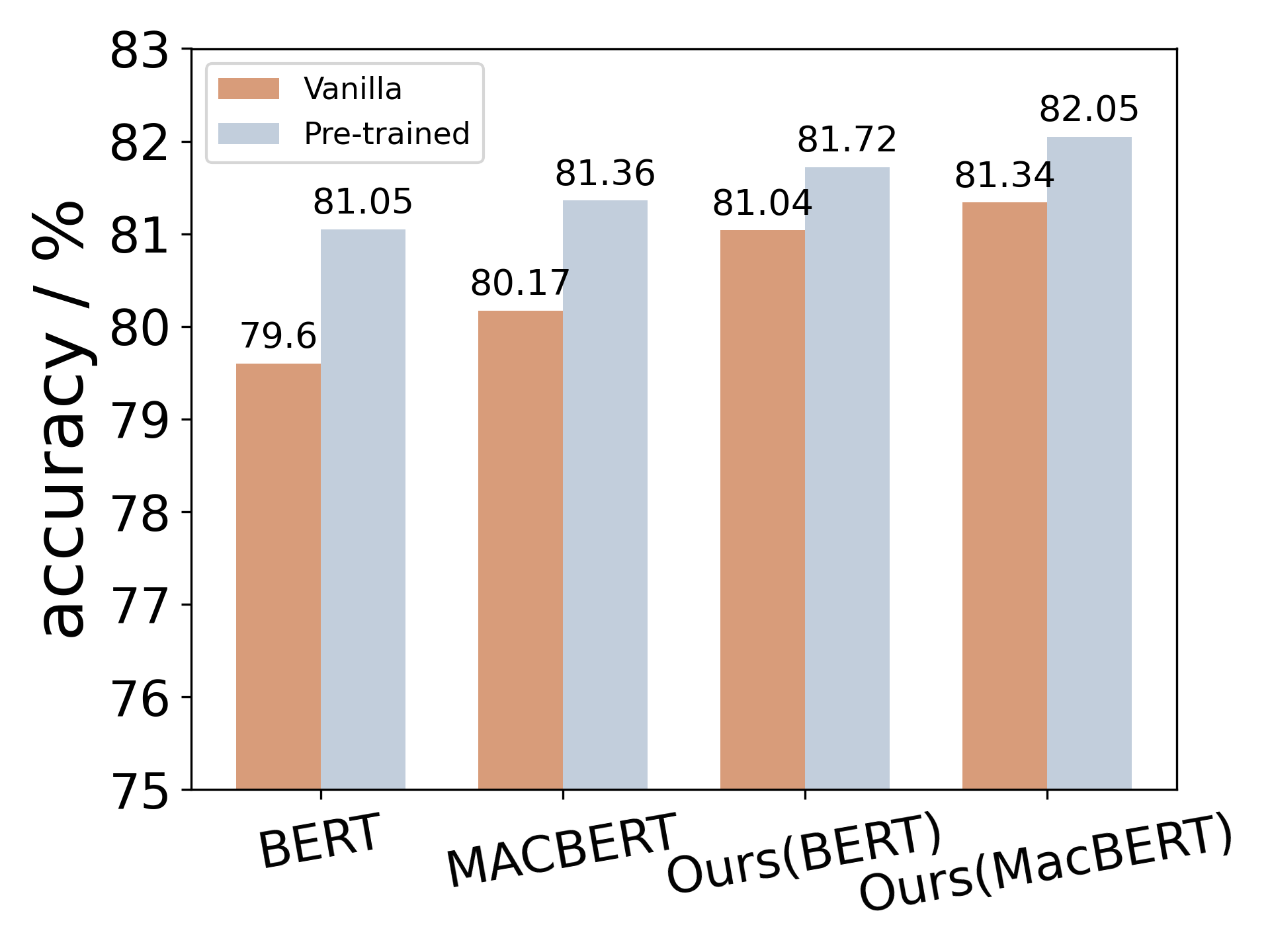}
    \label{fig:pt_subfigure1}
}
\hfill
\subfigure[Fine-grained]{
    \includegraphics[width=0.22\textwidth]{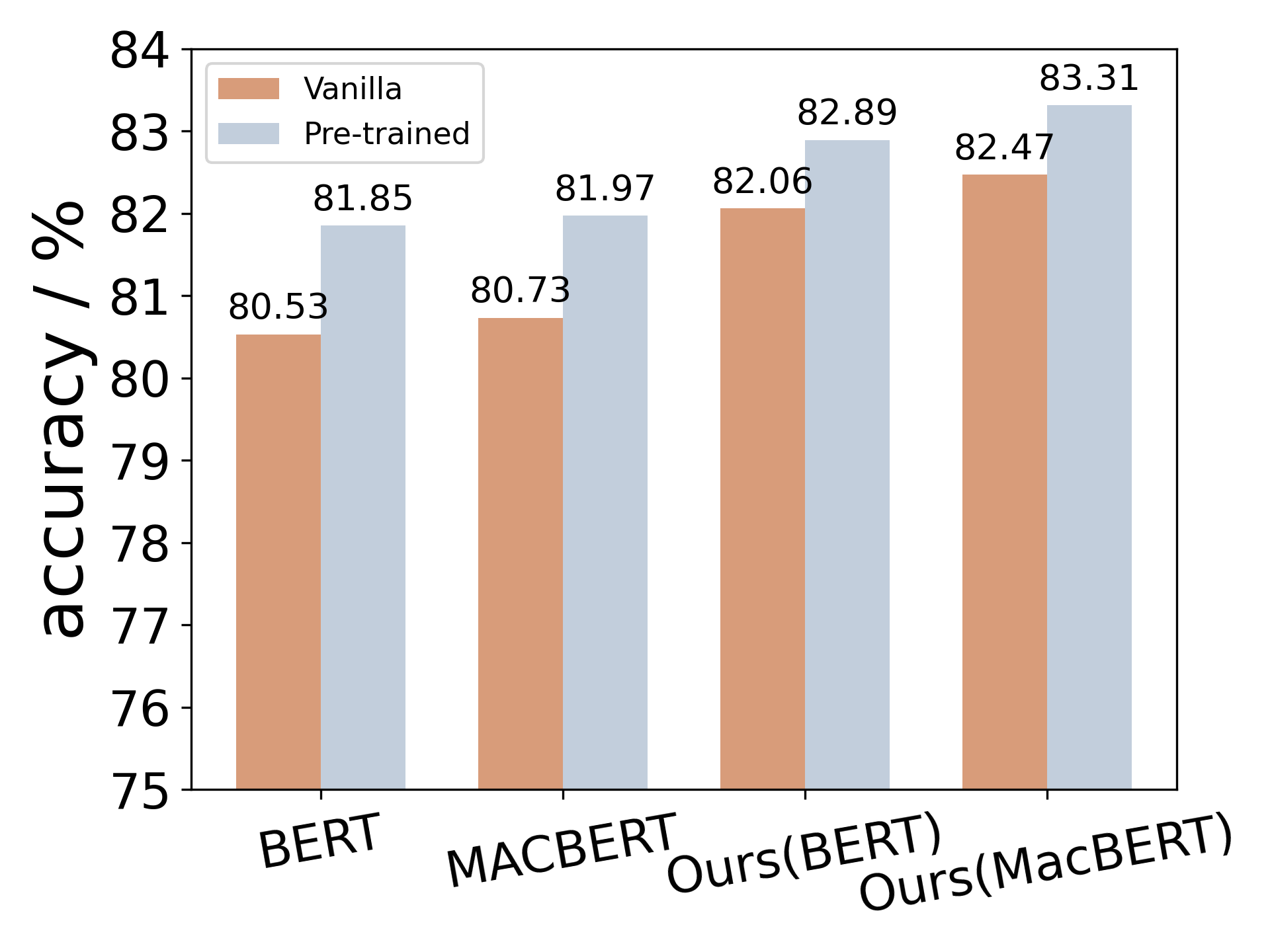}
    \label{fig:pt_subfigure2}
}
\vspace{-0.2cm}
\caption{Results of Domain Information Injection.}
\label{fig: domainresults}
\vspace{-0.2cm}
\end{figure}

To investigate the differences before and after domain-specific re-training of different pre-trained models, we designed comparative experiments to evaluate model performance under different conditions. We used the original BERT and MacBERT models, as well as the proposed models utilizing these pre-trained models, comparing their performance before and after re-training. The results are shown in Figure \ref{fig: domainresults}.

\vspace{-0.3cm}

With the same pre-trained models and classification network structures, re-training the language model can improve the classification accuracy. This is because the re-training data is related to Chinese medical information, allowing the language model to learn the intrinsic knowledge of medical texts, resulting in better performance of medical texts.

From this set of comparative data, it can be seen that using re-trained language models significantly improves accuracy and F1 scores. The proposed model, combined with the re-trained language models, achieves even better classification results. This indicates that re-training the language model with domain-specific data enhances its representation capability for domain-specific tasks. In this experiment, the language model re-trained with medical knowledge and Q\&A text data demonstrates better representation capability for texts that integrate colloquial expressions and medical knowledge.

\subsubsection{Parameters Study}
We also investigated the impact of different numbers of layers of dendritic neural network and LSTM layers on classification results to select the optimal algorithm model structure.

\begin{figure}[htbp]
\centering
\subfigure[DD layer]{
    \includegraphics[width=0.22\textwidth]{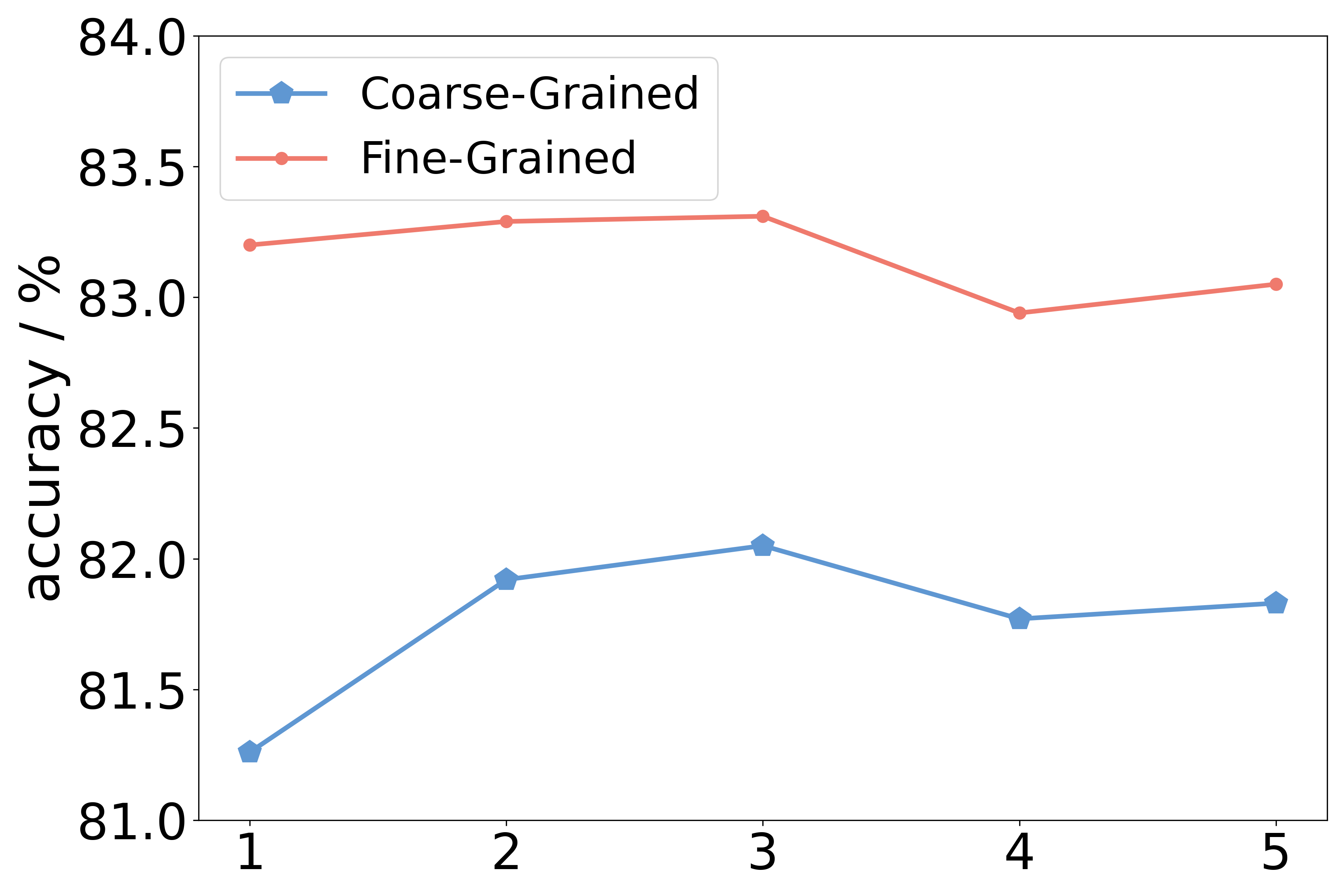}
    \label{fig:DD_subfigure1}
}
\hfill
\subfigure[LSTM layer]{
    \includegraphics[width=0.22\textwidth]{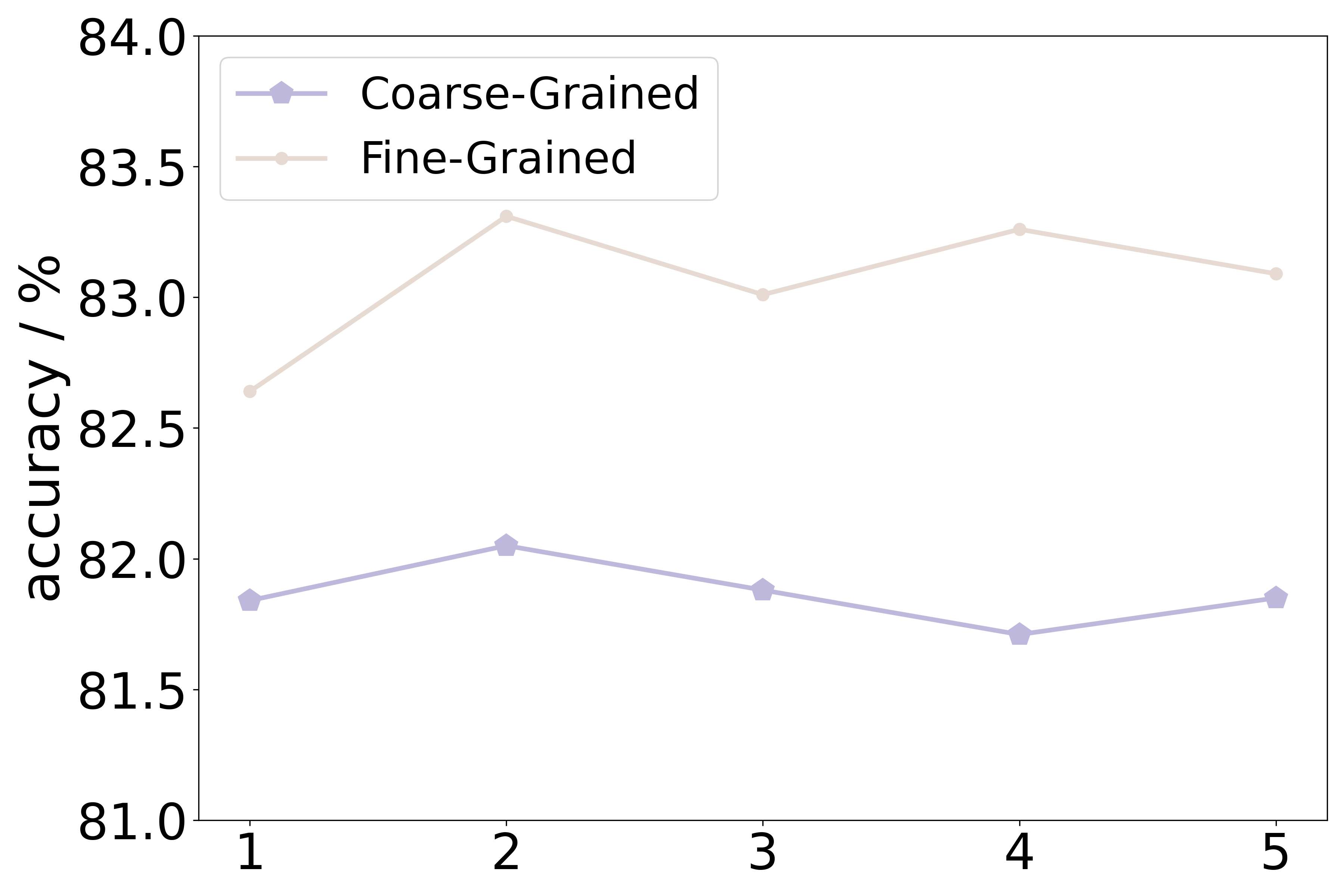}
    \label{fig:DD_subfigure2}
}
\vspace{-0.2cm}
\caption{Results of DD and LSTM layer Numbers.}
\label{fig: pararesults}
\vspace{-0.2cm}
\end{figure}

According to Figure \ref{fig: pararesults}, the dendritic neural network achieves the best classification performance with 3 layers, and the LSTM network achieves optimal performance with 2 layers.

\subsection{Prompting Triage System Experiment}

Prompt learning quickly completes tasks using pre-trained knowledge, even with limited data. To leverage this advantage, we created small sample datasets by removing high-data categories from the coarse-grained and fine-grained datasets, resulting in 73,604 entries across nine categories and 24,940 entries across 83 categories, respectively.

\begin{table}[h]
\caption{Prompt Learning Classification Results}
\vspace{-0.6cm}
\begin{center}
\resizebox{\linewidth}{!}{%
\begin{tabular}{llccc}
\toprule
\multirow{2}{*}{\textbf{Dataset}} & \multirow{2}{*}{\textbf{Method}} & \multicolumn{3}{c}{\textbf{Model}} \\
\cmidrule{3-5}
 &  & \textbf{Original BERT} & \textbf{Re-trained BERT} & \textbf{Supervised Learning} \\
\midrule
\multirow{2}{*}{\textbf{Coarse-grained}} & Original & 77.47\% & 79.39\% & 79.60\% \\
 & Small Sample & 86.36\% & 88.18\% & 87.16\% \\
\midrule
\multirow{2}{*}{\textbf{Fine-grained}} & Original & 79.30\% & 81.73\% & 83.31\% \\
 & Small Sample & 79.28\% & 80.11\% & 77.31\% \\
\bottomrule
\end{tabular}
}
\label{tab:prompt_results}
\end{center}
\vspace{-0.4cm}
\end{table}

Table \ref{tab:prompt_results} shows that the BERT model re-trained with collected data performs better in prompt learning-based classification, surpassing supervised learning on small datasets.

However, prompt learning has drawbacks: it requires more time due to predicting multiple tokens, and its performance can be slightly worse than supervised learning. This is due to the need to correctly predict all token positions and handle varying label lengths with padding symbols, which can interfere with pre-trained knowledge and reduce accuracy. Additionally, prompt learning is less effective with large datasets.

\subsection{Medical Consultation System Experiment}

\subsubsection{Overall Performance}
First, we compared the proposed framework's text generation model with a deep learning-based sequence-to-sequence (seq2seq) model. The seq2seq model is constructed with LSTM networks for both the Encoder and Decoder and includes information interaction methods such as attention mechanisms and information fusion representations. This model serves as a representative of deep learning methods that do not incorporate background information.

Additionally, we compared the proposed text generation model with the BART model. BART combines features of both autoencoding and autoregressive language models and is widely used in the field of text generation, especially for text summarization. In this study, we utilized the general Chinese BART model to build a dialogue system and trained it using medical dialogue data.

\begin{table}[htbp]
\caption{Comparison of Text Generation Results}
\vspace{-0.4cm}
\begin{center}
\resizebox{0.9\linewidth}{!}{
\begin{tabular}{lccc}
\toprule
\textbf{Evaluation Metric} & \textbf{Seq2Seq Model} & \textbf{BART} & \textbf{Our Model} \\
\midrule
Weight F1 $\uparrow$ & 0.1275 & 0.2070 & \textbf{0.2087} \\
Weight P $\uparrow$ & 0.1569 & \textbf{0.2215} & 0.2214 \\
Weight R $\uparrow$ & 0.1321 & 0.1998 & \textbf{0.2102} \\
BLEU-1~\cite{papineni2002bleu} $\uparrow$ & 0.0922 & \textbf{0.1214} & 0.1211 \\
CHRF~\cite{popovic2015chrf} $\uparrow$ & 0.0374 & 0.0769 & \textbf{0.0781} \\
GLEU~\cite{mutton2007gleu} $\uparrow$ & 0.0448 & 0.0977 & \textbf{0.1029} \\
NIST~\cite{doddington2002automatic} $\uparrow$ & 0.5398 & 1.3932 & \textbf{1.4049} \\
RIBES~\cite{isozaki2010automatic} $\uparrow$ & 0.0637 & 0.0827 & \textbf{0.0903} \\
TER~\cite{snover2006study} $\downarrow$ & 1.3519 & 1.0489 & \textbf{0.9997} \\
WMD~\cite{kusner2015word} $\uparrow$ & 0.5168 & 0.6192 & \textbf{0.6461} \\
BERT F1~\cite{zhang2019bertscore} $\uparrow$ & 0.568 & 0.646 & \textbf{0.648} \\
BERT P $\uparrow$ & 0.576 & 0.649 & \textbf{0.652} \\
BERT R $\uparrow$ & 0.562 & 0.644 & 0.644 \\
Information Entropy $\uparrow$ & 5.4215 & \textbf{5.7422} & 5.7291 \\
Lexical Diversity $\uparrow$ & 0.018 & 0.149 & \textbf{0.188} \\
KL Divergence -- & 0.7888 & 0.0549 & 0.0619 \\
Self BLEU Bigram $\downarrow$ & 0.9804 & \textbf{0.9875} & 0.9870 \\
Self BLEU Trigram $\downarrow$ & 0.9581 & \textbf{0.9694} & 0.9683 \\
Web Bert Similarity $\uparrow$ & 2.2324 & 2.2858 & \textbf{2.2960} \\
\bottomrule
\end{tabular}
}
\end{center}
\label{tab:comparison_text_generation}
\vspace{-0.4cm}
\end{table}

According to Table \ref{tab:comparison_text_generation}, the proposed model outperforms the traditional deep learning sequence-to-sequence (seq2seq) algorithm across almost all evaluation metrics. TER (Translation Edit Rate) measures the number of edits required to change a generated sentence to match the reference sentence, indicating the proximity between the generated and real texts. Therefore, a lower TER value implies a closer match between the generated and real texts. The superior performance of the model on these evaluation metrics is attributed to the knowledge embedded in the large model and the rationality of the proposed framework.

The Self-BLEU metric measures the BLEU score of a generated text against itself. A lower Self-BLEU score indicates greater diversity in the generated text, suggesting that the model produces more varied and information-rich content. However, it was observed that the seq2seq model had a lower Self-BLEU score than the proposed model. This is because the seq2seq model generates texts with many uncommon characters, increasing the perceived richness of the generated text. However, these characters do not add actual meaningful content, as reflected in the other metrics.

\subsubsection{Ablation study}
Next, we conducted ablation experiments to determine the importance of each component of the medical consultation system. Three comparative models were set up: a model without knowledge internalization, a model without input supplementation, and a model without both components.

\begin{table}[htbp]
\caption{Results of Text Generation Ablation Experiments}
\vspace{-0.6cm}
\begin{center}
\resizebox{\linewidth}{!}{%
\begin{tabular}{lcccc}
\toprule
\textbf{Evaluation Metric} & \textbf{Ours} & \textbf{w/o Knowledge} & \textbf{w/o Input} & \textbf{Neither} \\
\midrule
Weight F1 $\uparrow$ & \textbf{0.2087} & 0.2012 & 0.2005 & 0.1998 \\
Weight P $\uparrow$ & \textbf{0.2214} & 0.2108 & 0.2148 & 0.2111 \\
Weight R $\uparrow$ & \textbf{0.2102} & 0.2055 & 0.2016 & 0.2014 \\
BLEU-1 $\uparrow$ & \textbf{0.1211} & 0.1132 & 0.1181 & 0.1176 \\
CHRF $\uparrow$ & \textbf{0.0781} & 0.0745 & 0.0749 & 0.0736 \\
GLEU $\uparrow$ & \textbf{0.1029} & 0.0998 & 0.0989 & 0.0974 \\
NIST $\uparrow$ & \textbf{1.4049} & 1.3794 & 1.3755 & 1.3736 \\
RIBES $\uparrow$ & \textbf{0.0903} & 0.0850 & 0.0881 & 0.8237 \\
TER $\downarrow$ & \textbf{0.9997} & 1.0196 & 1.0215 & 1.0588 \\
WMD $\uparrow$ & \textbf{0.6461} & 0.6343 & 0.6351 & 0.6332 \\
BERT F1 $\uparrow$ & \textbf{0.648} & 0.647 & 0.645 & 0.644 \\
BERT P $\uparrow$ & \textbf{0.652} & \textbf{0.652} & 0.651 & 0.649 \\
BERT R $\uparrow$ & \textbf{0.644} & 0.643 & 0.641 & 0.641 \\
Information Density $\uparrow$ & \textbf{5.7291} & 5.6773 & 5.6226 & 5.6176 \\
Lexical Diversity $\uparrow$ & \textbf{0.188} & 0.168 & 0.155 & 0.156 \\
KL Divergence -- & 0.0619 & 0.0784 & 0.0973 & 0.0719 \\
Self BLEU Bigram $\downarrow$ & \textbf{0.9870} & 0.9888 & 0.9896 & 0.9898 \\
Self BLEU Trigram $\downarrow$ & \textbf{0.9683} & 0.9696 & 0.9716 & 0.9729 \\
Web Bert Similarity $\uparrow$ & \textbf{2.2960} & 2.2855 & 2.2819 & 2.2633 \\
\bottomrule
\end{tabular}
}
\label{tab:ablation_experiment_results}
\end{center}
\vspace{-0.4cm}
\end{table}

The results in Table \ref{tab:ablation_experiment_results} demonstrate that each step in the proposed framework is crucial. Removing any step results in a decline in the model's performance across various evaluation metrics, highlighting the superiority and rationality of the proposed structure.

The knowledge internalization step enables the generative pre-trained language model used for constructing the Q\&A model to learn more medical knowledge. This is similar to how human doctors leverage their background knowledge to provide answers. Hence, the knowledge internalization step enriches the Q\&A model with medical prior knowledge.

The input supplementation step simulates the process of human doctors searching for knowledge when encountering specific cases. By targeted acquisition of prior knowledge, combined with the powerful information retrieval capabilities of the knowledge graph, the model can generate more accurate response sentences. This process also contributes to the interpretability of the model construction process.

\section{Conclusion}




We collected and organized Chinese medical data, including datasets for automated medical triage and consultation, forming the foundation of our work.
Based on these datasets, we developed two systems: an intelligent triage system using text classification and a medical consultation system using a generative language model. Experimental results validated the effectiveness of both systems, demonstrating improved performance in medical diagnosis and consultation tasks.

Looking ahead, future work focuses on expanding datasets to reduce bias, enhancing intelligent triage systems with background knowledge and multi-modal methods, and refining medical consultation systems through alignment between LLM outputs and human preferences. At the same time, inspired by the field of recommendation systems~\cite{bai2023gorec, cai2024mitigating}, further considering rich user information combined with the powerful generalization ability of LLM to achieve personalized triage is also a future direction. The potential for further applications of LLMs is vast, enabling more flexible and realistic solutions in various fields.




\bibliographystyle{IEEEtran}
\bibliography{custom}

\end{document}